\documentclass{article}
\usepackage{ijcai05}
\usepackage{times}
\usepackage{graphicx}

\usepackage{amsmath}
\usepackage{amsfonts}

\usepackage{mathrsfs} % for mathscr
\usepackage{scalefnt}

\usepackage{url}

\newtheorem{definition}{Definition}
\newtheorem{theorem}{Theorem}

\title{A Simple Model to Generate Hard Satisfiable Instances}

\author{Ke Xu$^1$ \hspace{0.8cm} Fr\'ed\'eric Boussemart$^2$ \hspace{0.8cm} Fred Hemery$^2$ \hspace{0.8cm} Christophe Lecoutre$^2$\\
\begin{tabular}{c c c}
& \hspace{1cm} & \\
$^1$NLSDE Lab, School of Computers  & &         $^2$CRIL-CNRS FRE 2499,\\
Beihang University        &  &              Universit\'e d'Artois\\
Beijing 100083, China      &  &            Lens, France\\
kexu@nldse.buaa.edu.cn  &  & $\langle$lastname$\rangle$@cril.univ-artois.fr\\
\end{tabular}
}

\begin{document}

\maketitle

\begin{abstract}
  
In this paper, we try to further demonstrate that the models of random CSP instances proposed by \cite{XL_RB,XL_RD} are of theoretical and practical interest.
Indeed, these models, called RB and RD, present several nice features.
First, it is quite easy to generate random instances of any arity since no particular structure has to be integrated, or property enforced, in
such instances.
Then, the existence of an asymptotic phase transition can be guaranteed while applying a limited restriction on domain size and on constraint tightness. % (up to $50\%$ for binary constraints).
In that case, a threshold point can be precisely located and all instances have the guarantee to be hard at the threshold, i.e., to have an exponential tree-resolution complexity.
Next, a formal analysis shows that it is possible to generate forced satisfiable instances whose hardness is similar to unforced satisfiable ones.
This analysis is supported by some representative results taken from an intensive experimentation that we have carried out, using complete and incomplete search methods.

\end{abstract}

\section{Introduction}

Over the past ten years, the study of phase transition phenomena has been one of the most exciting areas in Computer Science and Artificial Intelligence.
Numerous studies have established that for many NP-complete problems (e.g., SAT and CSP), the hardest random instances occur, while a control parameter is varied accordingly, between an under-constrained region where all instances are almost surely satisfiable and an over-constrained region where all instances are almost surely unsatisfiable.
In the transition region, there is a threshold where half the instances are satisfiable. % and half the instances are unsatisfiable.
Generating hard instances is important both for understanding the complexity of the problems and for providing challenging benchmarks \cite{CM_finding}.

%Random instances generated at the threshold can be used to evaluate any complete search algorithm.
% but, without a guarantee about the satisfiability of the generated instances 
%However, if there is no guarantee about the satisfiability of the generated instances, one cannot use them to evaluate incomplete search algorithms.
%ndeed, if an incomplete search algorithm does not find a solution, it is impossible to conclude that the instance is unsatisfiable.
%One current issue is to produce hard instances 

Another remarkable progress in Artificial Intelligence has been the development of incomplete algorithms for various kinds of problems.
%The practical evaluation of such algorithms relies on the availability of hard instances whose satisfiability is guaranteed.
And, since this progress, one important issue has been to produce hard satisfiable instances in order to evaluate the efficiency of such algorithms, as the approach that involves exploiting a complete algorithm in order to keep random satisfiable instances generated at the threshold can only be used for instances of limited size.
Also, it has been shown that generating hard (forced) satisfiable instances is related to some open problems in cryptography such as computing a one-way function \cite{ILL_pseudo,CM_finding}. 

%To summarize, one current issue is to propose a method that allows producing hard instances as well as hard satisfiable instances.

In this paper, we mainly focus on random CSP (Constraint Satisfaction Problem) instances.
Initially, four ``standard'' models, denoted A, B, C and D \cite{SD_PT,G_random},  have been introduced to generate random binary CSP instances.
However, \cite{A_random} have identified a shortcoming of all these models. 
Indeed, they prove that random instances generated using these models suffer from (trivial) unsatisfiability as the number of variables increases.
To overcome the deficiency of these standard models, several alternatives have been proposed. 

On the one hand, \cite{A_random} have proposed a model E and \cite{M_models} a generalized model.
However, model E does not permit to tune the density of the instances and the generalized model requires an awkward exploitation of probability distributions.
Also, other alternatives correspond to incorporating some ``structure'' in the generated random instances.
Roughly speaking, it involves ensuring that the generated instances be arc consistent \cite{G_random} or path consistent \cite{GC_consistency}.
The main drawback of all these approaches is that generating random instances is no more quite a natural and easy task.

On the other hand, \cite{XL_RB,XL_RD}, \cite{FM_satisfiability} and \cite{S_constructing} have revisited standard models by controlling the way parameters change as the problem size increases.
The alternative model D scheme of \cite{S_constructing} guarantees the occurrence of a phase transition when some parameters are controlled and when the constraint tightness is within a certain range.
The two revised models, called RB and RD, of \cite{XL_RB,XL_RD} provide the same guarantee by varying one of two control parameters around a critical value that, in addition, can be computed.
Also, \cite{FM_satisfiability} identify a range of suitable parameter settings in order to exhibit a non-trivial threshold of satisfiability.
Their theoretical results apply to binary instances taken from model A and to ``symmetric'' binary instances from a so-called model B which, not corresponding to the standard one, associates the same relation with every constraint.

%We believe that
The models RB and RD present several nice features:
\begin{itemize}
\item it is quite easy to generate random instances of any arity as no particular structure has to be integrated, or property enforced, in such instances.
\item the existence of an asymptotic phase transition can be guaranteed while applying a limited restriction on domain size and on constraint tightness.
For instances involving constraints of arity $k$, the domain size is required to be greater than the k$^{th}$ root of the number of variables and the (threshold value of the) constraint tightness is required to be at most $\frac{k-1}{k}$.
%Tightness is then up to $50\%$ for binary constraints and $\approx 66\%$ for ternary constraints.
\item when the asymptotic phase transition exists, a threshold point can be precisely located, and all instances generated following models RB and RD have the guarantee to be hard at the threshold, i.e., to have an exponential tree-resolution complexity. %, contradicting the statement of \cite{GC_consistency} about the requirement of an extremely low tightness for all existing random models in order to have non-trivial threshold behaviors and guaranteed hard instances at the threshold,
\item it is possible to generate forced satisfiable instances whose hardness is similar to unforced satisfiable ones.
\end{itemize}

This paper is organized as follows.
After introducing models RB and RD, as well as some theoretical results (Section \ref{sec:the}), we provide a formal analysis about generating both forced and unforced hard satisfiable instances (Section  \ref{sec:generating}).
Then, we present the results of a large series of experiments that we have conducted (Section \ref{sec:experimental}), and, before concluding, we discuss some related work (Section \ref{sec:related}).

\section{Theoretical background \label{sec:the}}

%%\begin{definition}
A constraint network consists of a finite set of variables such that each variable $X$ has an associated domain denoting the set of values allowed for $X$, and a a finite set of constraints such that each constraint $C$ has an associated relation denoting the set of tuples allowed for the variables involved in $C$. 
%\end{definition}

A solution is an assignment of values to all the variables such that all the constraints are satisfied. 
A constraint network is said to be satisfiable (sat, for short) if it admits at least a solution.
The Constraint Satisfaction Problem (CSP), whose task is to determine if a given constraint network, also called CSP instance, is satisfiable, is NP-complete.
%A constraint network is also called CSP instance.
%In this paper, solving a CSP instance involves either finding one solution or determining its unsatisfiability.

In this section, we introduce some theoretical results taken from \cite{XL_RB,XL_RD}.
First, we introduce a model, denoted RB, that represents an alternative to model B. % (e.g., see \cite{G_random}).
Note that, unlike model B, model RB allows selecting constraints with repetition.
But the main difference of model RB with respect to model B is that the domain size of each variable grows polynomially with the number of variables.

\begin{definition}[Model RB]
A class of random CSP instances of model RB is denoted RB($k$,$n$,$\alpha$,$r$,$p$) where:
\begin{itemize}
\item $k \geq 2$ denotes the arity of each constraint,
\item $n$ denotes the number of variables,
\item $\alpha > 0$ determines the domain size $d = n^{\alpha}$ of each variable,
\item $r > 0$ determines the number $m = r.n.\ln n$ of constraints,
\item $1 > p > 0$ denotes the tightness of each constraint.
\end{itemize}
To build one instance $P \in$ RB($k$,$n$,$\alpha$,$r$,$p$), we select with repetition $m$ constraints, each one formed by selecting $k$ distinct variables and $p.d^k$ distinct unallowed tuples (as $p$ denotes a proportion).
\end{definition}

When fixed, $\alpha$ and $r$ give an indication about the growth of the domain sizes and of the number of constraints as $n$ increases since $d = n^{\alpha}$ and $m=rn \ln n$, respectively.
It is then possible, for example, to determine the critical value $p_{cr}$ of $p$ where the hardest instances must occur.
Indeed, we have $p_{cr} = 1 - e^{-\alpha/r}$ which is equivalent to the expression of $p_{cr}$ given by \cite{SD_PT}.

Another model, denoted model RD, is similar to model RB except that $p$ denotes a probability instead of a proportion.
For convenience, in this paper, we will exclusively refer to model RB although all given results hold for both models.

In \cite{XL_RB}, it is proved that model RB, under certain conditions, not only avoids trivial asymptotic behaviors but also guarantees exact phase transitions.
More precisely, with Pr denoting a probability distribution, the following theorems hold.

\begin{theorem} \label{the:1}
If $k$, $\alpha > \frac{1}{k}$ and $p \leq \frac{k-1}{k}$ are constants then
\begin{align*}
\lim_{n \rightarrow \infty} Pr[P \in RB(k,n,\alpha,r,p)~ is ~ sat ~ ] =
\left \{ 
\begin{array}{cc}
1 & if~ r < r_{cr} \\
0 & if~ r > r_{cr}
\end{array}
\right.
\end{align*}
where $r_{cr} = -\frac{\alpha}{\ln(1-p)}$.
\end{theorem}

\begin{theorem} \label{the:2}
If $k$, $\alpha > \frac{1}{k}$ and $p_{cr} \leq \frac{k-1}{k}$ are constants then 
\begin{align*}
\lim_{n \rightarrow \infty} Pr[P \in RB(k,n,\alpha,r,p)~ is ~ sat ~ ] =
\left \{ 
\begin{array}{cc}
1 & if~ p < p_{cr} \\
0 & if~ p > p_{cr}
\end{array}
\right.
\end{align*}
where $p_{cr} = 1 - e^{-\frac{\alpha}{r}}$. % and Pr(Sat) denotes the probability that a random CSP instance generated following model RB is satisfiable.
\end{theorem}

Remark that the condition $p_{cr} \leq \frac{k-1}{k}$ is equivalent to $ke^{-\frac{\alpha}{r}} \geq 1$ given in \cite{XL_RB}.
%Table \ref{tab:limits} gives the limits of Theorems \ref{the:1} and \ref{the:2} for different arities.
Theorems \ref{the:1} and \ref{the:2} indicate that a phase transition is guaranteed provided that the domain size is not too small and the constraint tightness or the threshold value of the constraint tightness not too large.
As an illustration, for instances involving binary (resp. ternary) constraints, the domain size is required to be greater than the square (resp. cubic) root of the number of variables and the constraint tightness or threshold value of the tightness is required to be at most $50\%$ (resp. $\approx 66\%$).   

%\begin{table}[htb]
%\begin{scriptsize}
%\begin{center}
%\begin{tabular}{|c|r|r|}
%\hline
%$k$ & $\alpha$ & $p or p_{cr} $\\
%\hline
%$2$ & $> 1/2$ & $\leq 1/2$\\
% \hline
%$3$ & $> 1/3$ & $\leq 2/3$\\
%\hline
%$4$ & $> 1/4$ & $\leq 3/4$\\
%\hline
%\end{tabular}
%\caption{Limits of Theorem \ref{the:1} and \ref{the:2} \label{tab:limits} }
%\end{center}
%\end{scriptsize}
%\end{table}

%Theorems \ref{the:1} and \ref{the:2} establish that the instances above the threshold are almost surely unsatisfiable.

The following theorem establishes that unsatisfiable instances of model RB almost surely have the guarantee to be hard.
A similar result for model A has been obtained by \cite{FM_satisfiability} with respect to binary instances. % (i.e. k=$2$).
%When we say that a property holds almost surely, it means that this property holds with probability tending to $1$ as the number of variables tends to infinity.

\begin{theorem} \label{the:3}
If $P \in$ RB($k$,$n$,$\alpha$,$r$,$p$) and $k$, $\alpha$, $r$ and $p$ are constants, then, almost surely\footnote{We say that a property holds almost surely when this property holds with probability tending to $1$ as the number of variables tends to infinity.}, $P$ has no tree-like resolution of length less than  $2^{\Omega(n)}$.
\end{theorem}

%\begin{theorem} \label{the:3}
%\begin{align*}
%\lim_{n \rightarrow \infty} Pr[RES(P) = 2^{\Omega(n)}~ with ~ P \in RB(n,\alpha,k,r,p)] = 1
%\end{align*}
%where RES($P$) denotes the (tree-like) resolution complexity of the CNF formula corresponding to a SAT encoding of $P$.
%\end{theorem}

The proof, which is based on a strategy following some results of \cite{BW_short,M_resolution}, is omitted but can be found in \cite{XL_RD}.

%We have also established exponential lower bounds on the complexity of solving random satisfiable instances of model RB below the threshold. 
%Besides, such exponential lower bounds hold for forced satisfiable instances, i.e. instances on which a solution is imposed (see next Section).
%However, due to lack of space, these technical results are not inserted in this paper.
%So, we report the reader to Theorems $5$ and $6$ of Section $5$ in \cite{XL_RD}.

To summarize, model RB guarantees exact phase transitions and hard instances at the threshold. 
It then contradicts the statement of \cite{GC_consistency} about the requirement of an extremely low tightness for all existing random models in order to have non-trivial threshold behaviors and guaranteed hard instances at the threshold.

\section{Generating hard satisfiable instances \label{sec:generating}}

For CSP and SAT, there is a natural strategy to generate {\it forced} satisfiable instances, i.e., instances on which a solution is imposed.
It suffices to proceed as follows: first generate a random (total) assignment $t$ and then generate a random instance with $n$ variables and $m$ constraints (clauses for SAT) such that any constraint violating $t$ is rejected.
$t$ is then a {\it forced} solution.
This strategy, quite simple and easy to implement, allows generating hard forced satisfiable instances of model RB provided that Theorem $1$ or $2$ holds.
Nevertheless, this statement deserves a theoretical analysis. 

Assuming that $d$ denotes the domain size ($d$ = $2$ for SAT), we have exactly $d^{n}$ possible (total) assignments, denoted by $t_1,t_2,\cdots,t_{d^{n}}$, and $d^{2n}$ possible assignment pairs where an {\it assignment pair} $<t_i,t_j>$ is an ordered pair of two assignments $t_i$ and $t_j$.
We say that $<t_i,t_j>$ satisfies an instance if and only if both $t_i$ and $t_j$ satisfy the instance.
%Let $t_i$ be an assignment that has been forced to be a solution.
Then, the expected (mean) number of solutions $E_{f}[N]$ for instances that are forced to satisfy an assignment $t_i$ is: %(which is then a forced solution) is:
\[
E_{f}[N] = \overset{d^n}{\underset{j=1}{\sum}} \frac{\Pr[<t_i,t_j>]}{\Pr[<t_i,t_i>]}
\]
\noindent where $\Pr[<t_i,t_j>]$ denotes the probability that $<t_i,t_j>$ satisfies a random instance.
Note that $E_{f}[N]$ should be independent of the choice of the forced solution $t_i$.
So we have: 
\[
E_{f}[N]=\frac{\underset{1\leq i,j\leq d^{n}}{%
{\displaystyle\sum}
}\Pr[<t_{i},t_{j}>]}{d^{n}\Pr[<t_{i},t_{i}>]}=\frac{E[N^{2}]}{E[N]}.
\]
\noindent where $E[N^{2}]$ and $E[N]$ are, respectively, the second moment and the first moment of the number of solutions for random unforced instances.
%It is straightforward to derive, from the results on ordered pairs of assignments for random $k$-SAT in [34], that the expected number of solutions for random forced satisfiable instances is equal to $E(N^{2})/E(N),$ where $E(N^{2})$ and $E(N)$ are, respectively, the second moment and the first moment of the number of solutions for instances generated randomly. 

For random 3-SAT,  it is known that the strategy mentioned above is unsuitable as it produces a biased sampling of instances with many solutions clustered around $t$ \cite{A_generating}.
Experiments show that forced satisfiable instances are much easier to solve than unforced satisfiable instances.
In fact, it is not hard to show that, asymptotically, $E[N^{2}]$ is exponentially greater than $E^{2}[N]$. 
%The same conclusion can also be found in \cite{A_asymptotic}.
Thus, the expected number of solutions for forced satisfiable instances is exponentially larger than the one for unforced satisfiable instances.
It then gives a good theoretical explanation of why, for random 3-SAT, the strategy is highly biased towards generating instances with many solutions.

For model RB, recall that when the exact phase transitions were established \cite{XL_RB}, it was proved that $E[N^{2}]/E^{2}[N]$ is asymptotically equal to $1$ below the threshold, where almost all instances are satisfiable, i.e. $E[N^{2}]/E^{2}[N]\approx 1$ for $r<r_{cr}$ or $p<p_{cr}$.
Hence, the expected number of solutions for forced satisfiable instances below the threshold is asymptotically equal to the one for unforced satisfiable instances, i.e. $E_{f}[N]=E[N^{2}]/E[N]\approx E[N]$.
In other words, when using model RB, the strategy has almost no effect on the number of solutions and does not lead to a biased sampling of instances with many solutions. 

In addition to the analysis above, we can also study the influence of the strategy on the distribution of solutions with respect to the forced solution.
%we know, from $E(N^{2})\approx E^{2}(N),$ that for randomly 
%generated instances of RB/RD, the distribution of the number of solutions is 
%quite uniform with concentration around $E(N).$  Note that the truth assignment 
%$t$ is generated randomly and the strategy will in fact generate all the possible 
%satisfiable instances with $t$ as the satisfying assignment. 
We first define the distance $d^f(t_i,t_j)$ between two assignments $t_i$ and $t_j$ as the proportion of variables that have been assigned a different value in $t_i$ and $t_j$. 
We have $0\leq d^{f}(t_i,t_j)\leq 1$. 

%\binom{n}{n\delta}\left(  n^{\alpha}-1\right)  ^{n\delta}  

For forced satisfiable instances of model RB, with $E_f^{\delta}[N]$ denoting the expected number of solutions whose distance from the forced solution (identified as $t_i$, here) is equal to $\delta$, we obtain by an analysis similar to that in \cite{XL_RB}:
{\footnotesize 
\begin{align*}
& E_{f}^{\delta}[N] = \overset{d^n}{\underset{j=1}{\sum}} \frac{\Pr[<t_i,t_j>]}{\Pr[<t_i,t_i>]}\text{ \ \ with }d^{f}(t_i,t_j)=\delta\\
& =\binom{n}{n\delta} \left(n^{\alpha}{\text - }1\right)^{n\delta} \left[\frac{\binom{n-n\delta}{k}}{\binom{n}{k}}+(1-p)\left(  1-\frac{\binom{n-n\delta}{k}}{\binom{n}{k}}\right)  \right]  ^{rn\ln n}\\
& = \exp\left[ n\ln n\left(  r\ln\left(1-p+p(1-\delta)^{k}\right)  +\alpha \delta \right)+O(n)  \right]  .
\end{align*}
}
It can be shown, from the results in \cite{XL_RB} that $E_{f}^{\delta}[N]$, for $r<r_{cr}$ or $p<p_{cr}$, is asymptotically maximized when $\delta$ takes the largest possible value, i.e. $\delta=1$.

For unforced satisfiable instances of model RB, with $E^{\delta}[N]$ denoting the expected number of solutions whose distance from $t_i$ (not necessarily a solution) is equal to $\delta$, we have:
\begin{align*}
E^{\delta}[N]  & =\binom{n}{n\delta}\left(  n^{\alpha}-1\right)  ^{n\delta}\left(
1-p\right)  ^{rn\ln n}\\
& =  \exp\left[  n\ln n\left(  r\ln(1-p)+\alpha
\delta \right)+O(n)  \right]  .
\end{align*}
It is straightforward to see that the same pattern holds for this case, i.e. $E^{\delta}[N]$ is asymptotically maximized when $\delta=1$.

Intuitively, with model RB, both unforced satisfiable instances and instances forced to satisfy an assignment $t$ are such that most of their solutions distribute far from $t$.
This indicates that, for model RB, the strategy has little effect on the distribution of solutions, and is not biased towards generating instances with many solutions around the forced one.

For random 3-SAT, similarly, we can show that as $r$ (the ratio of clauses to variables) approaches $4.25$, $E_{f}^{\delta}[N]$ and $E^{\delta}[N]$ are asymptotically maximized when $\delta \approx 0.24$ and $\delta=0.5$, respectively.
This means, in contrast to model RB, that when $r$ is near the threshold, most solutions of forced instances distribute in a place much closer to the forced solution than solutions of unforced satisfiable instances.

\section{Experimental results \label{sec:experimental}}

As all introduced theoretical results hold when $n \rightarrow \infty$, the practical exploitation of these results is an issue that must be addressed.
In this section, we give some representative experimental results which indicate that practice meets theory even if the number $n$ of variables is small.
Note that different values of parameters $\alpha$ and $r$ have been selected in order to illustrate the broad spectrum of applicability of model RB.

\begin{figure}[h]
\centerline{\includegraphics[width=2.57in]{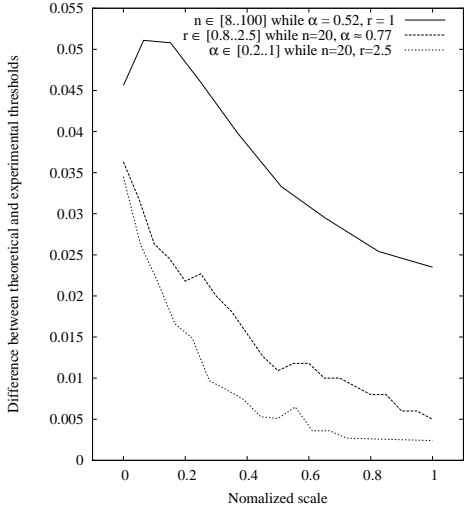}}
\caption{Difference between theoretical and experimental thresholds against $\alpha$, $r$ and $n$}%. %Precision of Theorem \ref{the:2} against different values of $\alpha$, $r$ and $n$} %Difference between theoretical and experimental thresholds for $n=20$, $r=2.5$, $k=2$ and  $d \in [2..50]$ (hence, $\alpha \in  [0.2,1.3]$)}
\label{fig:diff}
\end{figure}

%Theorems \ref{the:1} and \ref{the:2} guarantee asymptotically the presence of a phase transition and the exact localization of the threshold.
First, it is valuable to know in practice, to what extent, Theorems \ref{the:1} and \ref{the:2} give precise thresholds according to different values of $\alpha$, $r$ and $n$. 
The experiments that we have run wrt Theorem \ref{the:2}, as depicted in Figure \ref{fig:diff}, suggest that all other parameters being fixed, the greater the value of $\alpha$, $r$ or $n$ is, the more precise Theorem \ref{the:2} is.
More precisely, in Figure \ref{fig:diff}, the difference between the threshold theoretically located and the threshold experimentally determined is plotted against $\alpha \in  [0.2,1]$ ($d \in [2..20]$), against $r \in  [0.8,2.5]$ ($m \in [50..150]$) and against  $n \in [8..100]$.
Note that the vertical scale refers to the difference in constraint tightness and that the horizontal scale is normalized (value $0$ respectively corresponds to $n=8$, $\alpha=0.2$ and $r=0.8$, etc.).

\begin{figure}[bh]
\centerline{\includegraphics[width=2.57in]{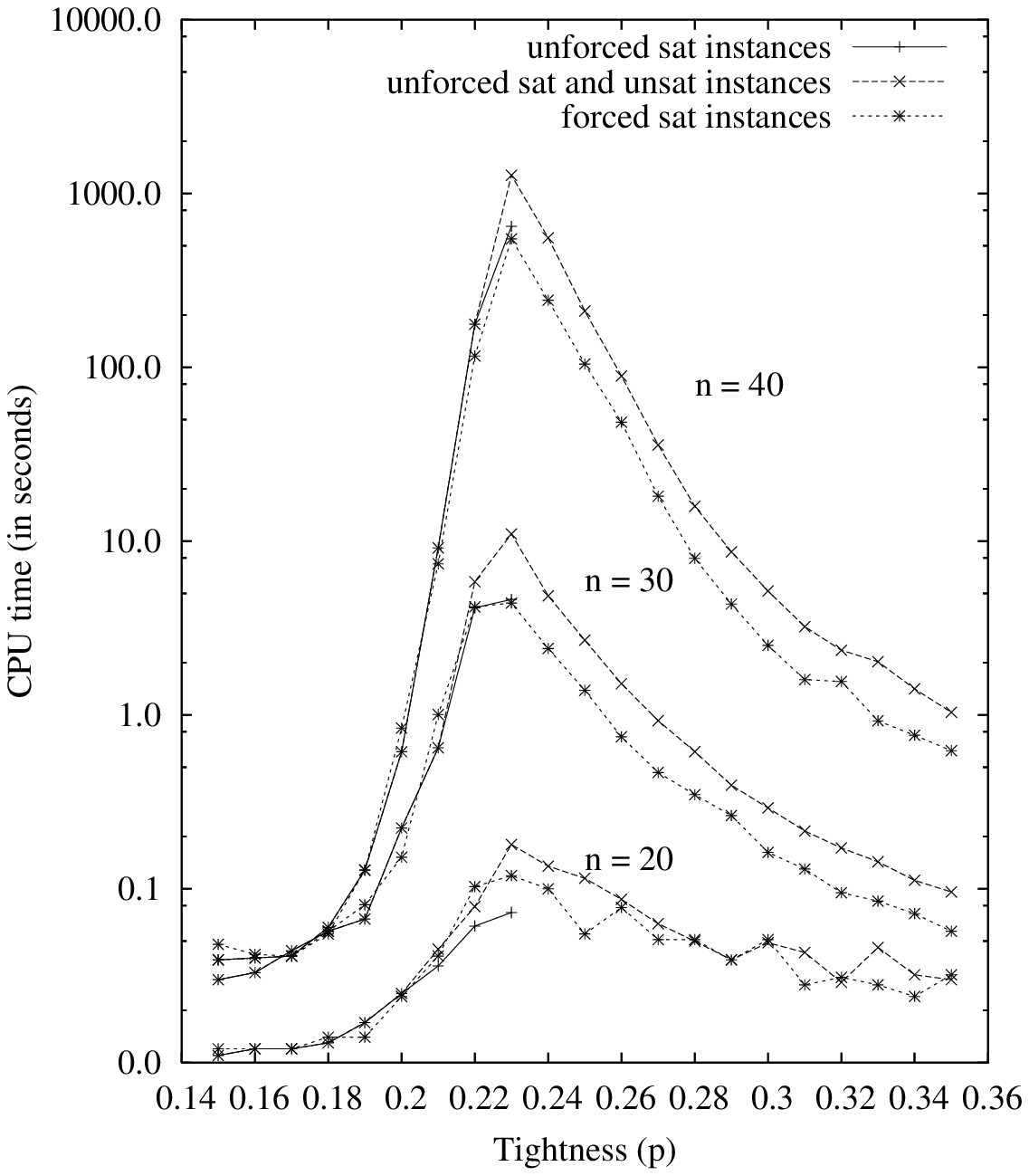}}
\caption{Mean search cost ($50$ instances) of solving instances in RB($2$,$\{20,30,40\}$,$0.8$,$3$,$p$)} %Observed phase transitions when $\alpha=0.8$, $r=3$, $k=2$ and $n \in \{20,30,40\}$}
\label{fig:phase1}
\end{figure}
%\vspace{-5mm}

To solve the random instances generated by model RB, we have used a systematic backtracking search algorithm (MAC) and a local search algorithm (tabu search).
Both algorithms have been equipped with a search heuristic that learns from conflicts \cite{BHLS_boosting}.

\begin{figure}[t]
\centerline{\includegraphics[width=2.57in]{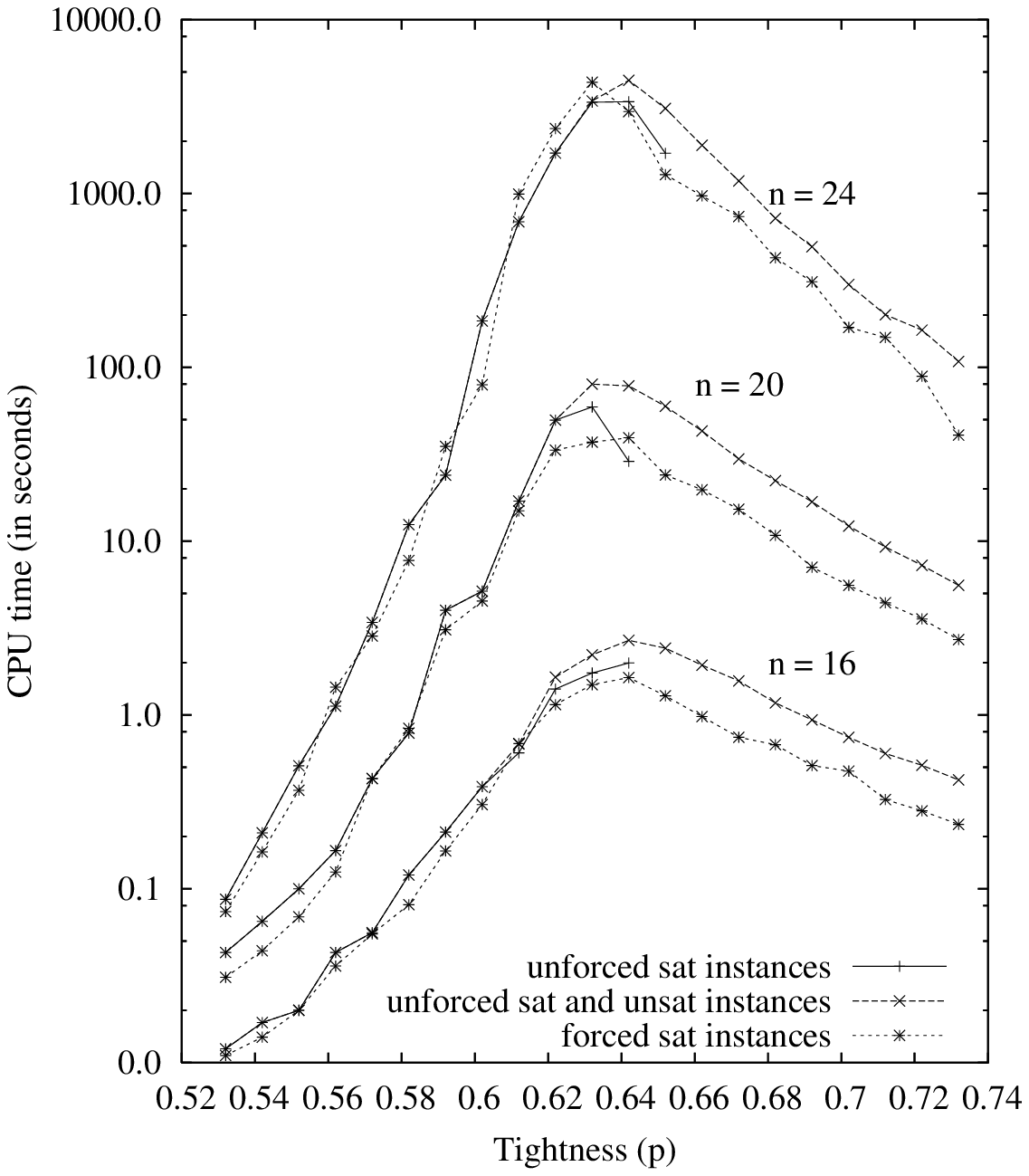}}
\caption{Mean search cost ($50$ instances) of solving instances in RB($3$,$\{16,20,24\}$,$1$,$1$,$p$)} %Observed phase transitions when $\alpha=0.8$, $r=3$, $k=2$ and $n \in \{20,30,40\}$}
\label{fig:phase2}
\end{figure}

We have studied the difficulty of solving with MAC the binary instances of model RB generated around the theoretical threshold $p_{cr} \approx 0.23$ given by Theorem \ref{the:2} for  $k=2$, $\alpha=0.8$, $r=3$ and $n \in \{20,30,40\}$.
In Figure \ref{fig:phase1}, it clearly appears that the hardest instances are located quite close to the theoretical threshold and that the difficulty grows exponentially with $n$ (note the use of a log scale).
It corresponds to a phase transition (not depicted here, due to lack of space).
A similar behavior is observed in Figure \ref{fig:phase2} with respect to ternary instances generated around the theoretical threshold $p_{cr} \approx 0.63$ for  $k=3$, $\alpha=1$, $r=1$ and $n \in \{16,20,24\}$.

As the number and the distribution of solutions are the two most important factors determining the cost of solving satisfiable instances, we can expect, from the analysis given in Section \ref{sec:generating}, that for model RB, the hardness of solving forced satisfiable instances should be similar to that of solving unforced satisfiable ones. 
This is what is observed in Figure \ref{fig:phase1}.

\begin{figure}[h]
\centerline{\includegraphics[width=2.57in]{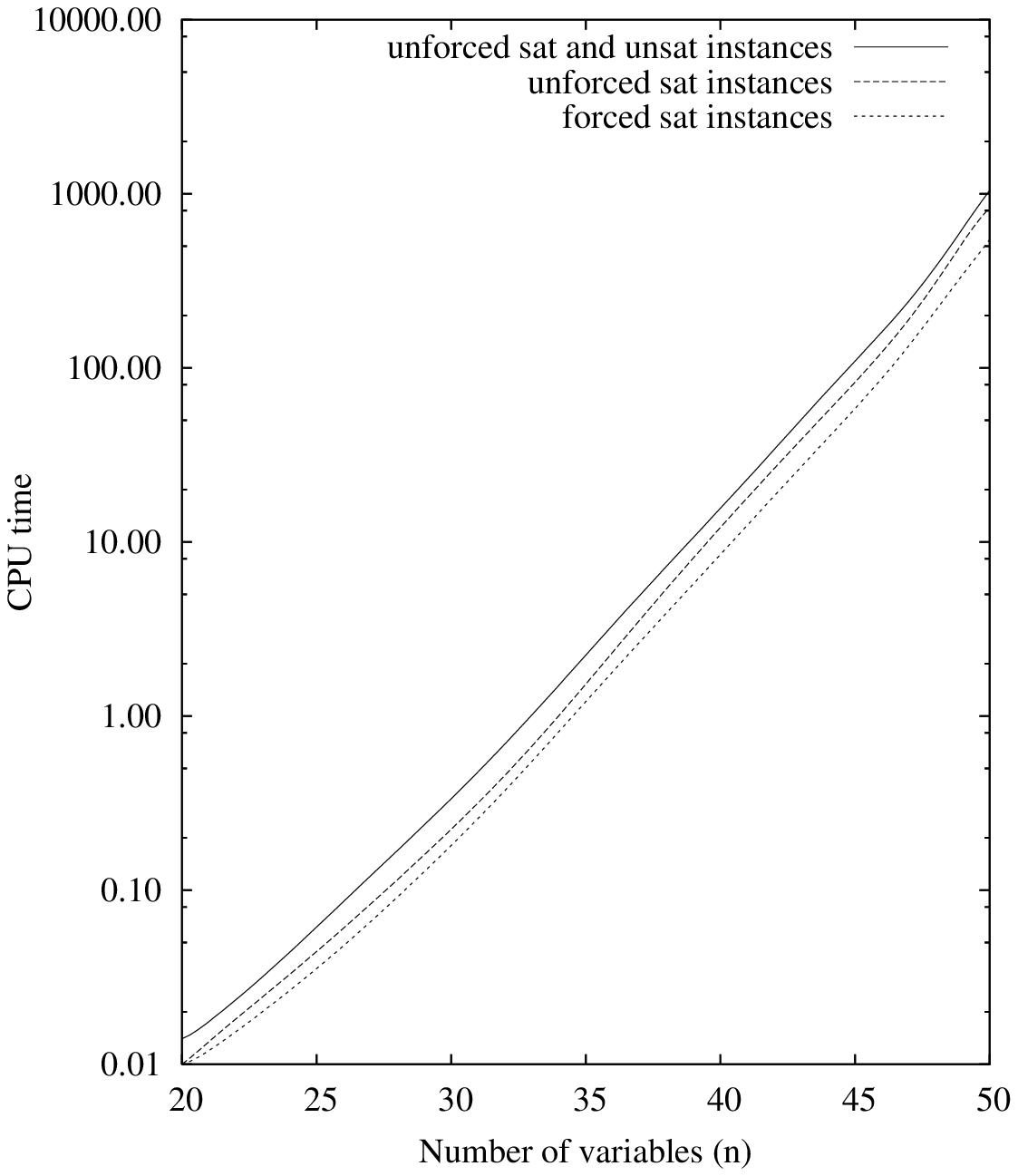}}
\caption{Mean search cost ($50$ instances) of solving instances in RB($2$,$[20..50]$,$0.8$,$1.5$,$p_{cr}-0.01$)} % when $\alpha=0.8$, $r=1.5$, $k=2$ and $n \in [20..70]$ at $p_{cr} \approx 0.41 - 0.2$}
\label{fig:forced}
\end{figure}

To confirm this, we have focused our attention to a point just below the threshold as we have then some (asymptotic) guarantee about the difficulty of both unforced and forced instances (see Theorems $5$ and $6$ in \cite{XL_RD}) and the possibility of generating easily unforced satisfiable instances.
Figure \ref{fig:forced} shows the difficulty of solving with MAC both forced and unforced instances of model RB at $p_{cr}-0.01 \approx 0.40$ for $k=2$, $\alpha=0.8$, $r=1.5$ and $n \in [20..50]$.
%The hardness of solving forced satisfiable instances is similar to that of solving unforced satisfiable ones

\begin{figure}[h]
\centerline{\includegraphics[width=2.57in]{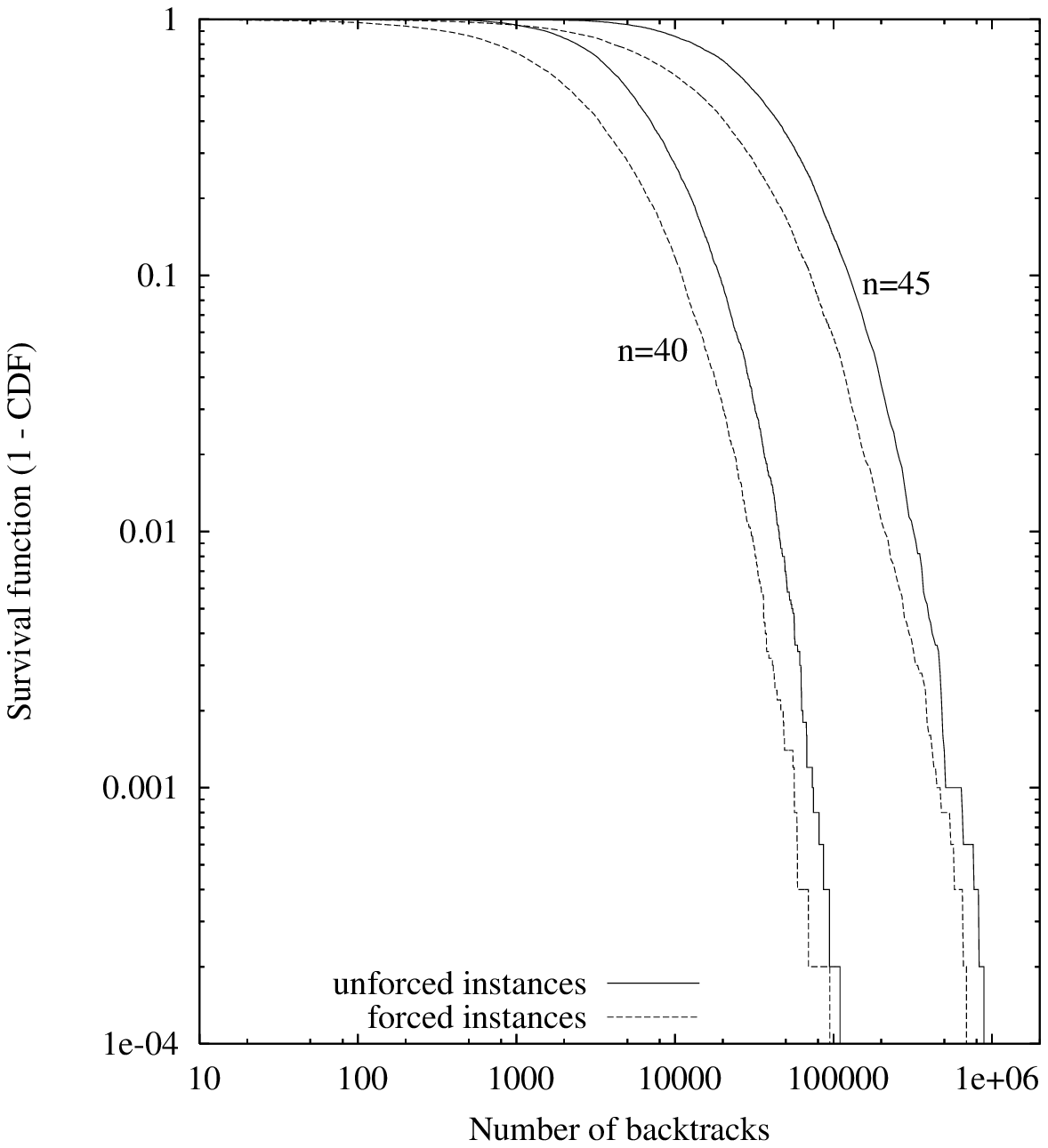}}
\caption{Non heavy-tailed regime for instances in RB($2$,$\{40,45\}$,$0.8$,$1.5$,$p_{cr} \approx 0.41$)}
\label{fig:heavyTail}
\end{figure}

To confirm the inherent difficulty of the (forced and unforced) instances generated at the threshold, we have also studied the runtime distribution produced by a randomized search algorithm on distinct instances \cite{GFSB_statistical}.
For each instance, we have performed $5000$ independent runs.
Figure \ref{fig:heavyTail} displays the survival function, which corresponds to the probability of a run taking more than $x$ backtracks, of a randomized MAC algorithm for two representative instances generated at $p_{cr} \approx 0.41$ for $k=2$, $\alpha=0.8$, $r=1.5$ and $n \in \{40,45\}$.
One can observe that the runtime distribution (a log-log scale is used) do not correspond to an heavy-tailed one, i.e., a distribution characterized by an extremely long tail with some infinite moment.
It means that all runs behave homogeneously and, therefore, it suggests that the instances are inherently hard \cite{GFSB_statistical}.

\begin{figure}[t]
\centerline{\includegraphics[width=2.57in]{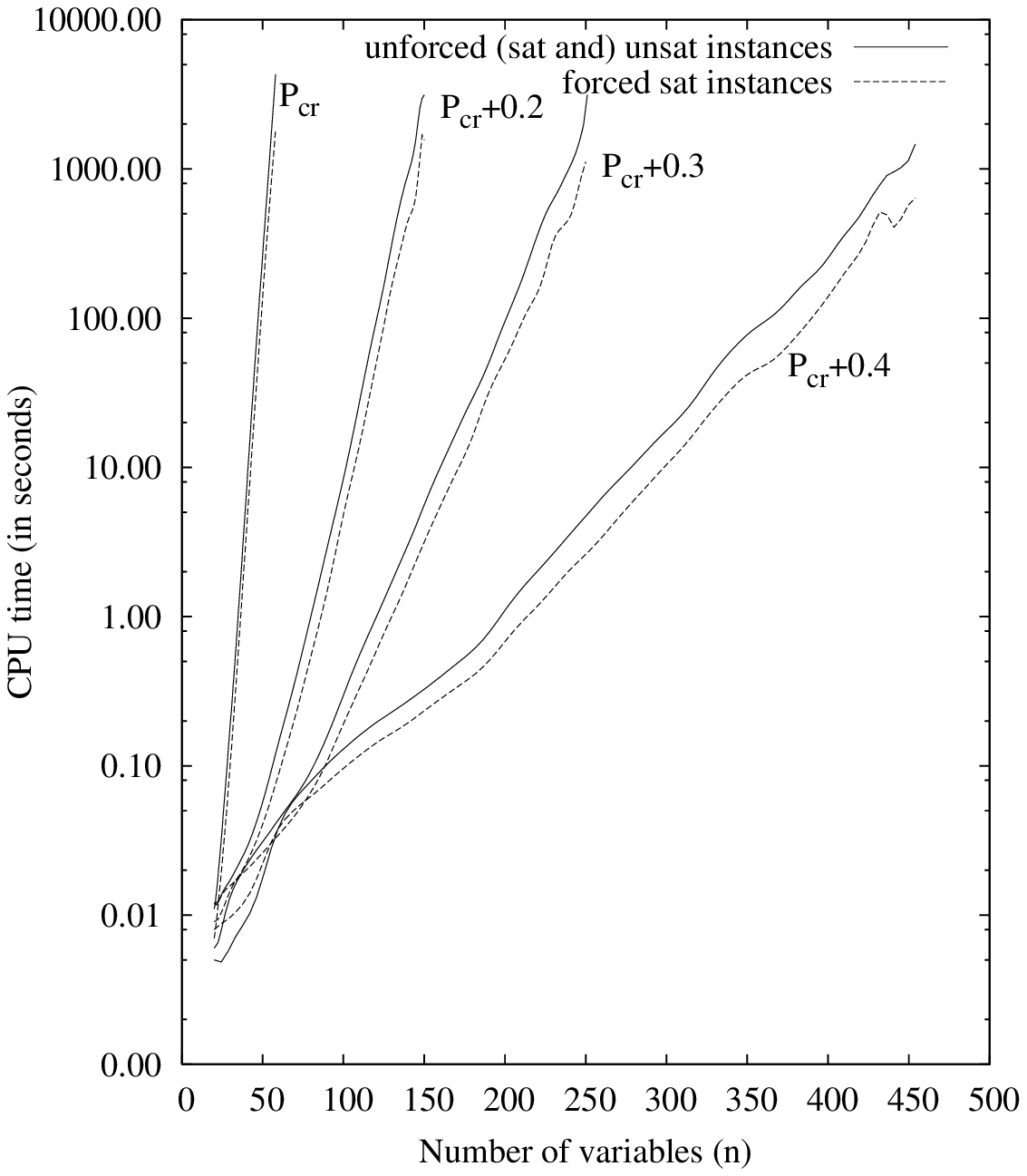}}
\caption{Mean search cost ($50$ instances) of solving instances in RB($2$,$[20..450]$,$0.8$,$1.5$,$p$)} % with $p \in \{p_{cr} \approx 0.41, p_{cr}+0.2,p_{cr}+0.3,p_{cr}+0.4\}$}
\label{fig:above}
\end{figure}

Then, we have focused on unforced unsatisfiable instances of model RB as Theorem \ref{the:3} indicates that such instances have an exponential resolution complexity. % (and also to experiment large random instances).
We have generated unforced and forced instances with different constraint tightness $p$ above the threshold $p_{cr} \approx 0.41$ for $k=2$, $\alpha=0.8$, $r=1.5$ and $n \in [20..450]$.
Figure \ref{fig:above} displays the search effort of a MAC algorithm to solve such instances against the number of variables $n$.
It is interesting to note that the search effort grows exponentially with $n$, even if the exponent decreases as the tightness increases.
%Also, although not currently supported by any theoretical result, it appears here that forced and unforced instances have a similar hardness.
Also, although not currently supported by any theoretical result (Theorems $5$ and $6$ of \cite{XL_RD} hold only for forced instances below the threshold) it appears here that forced and unforced instances have a similar hardness.

\begin{figure}[t]
\centerline{\includegraphics[width=2.57in]{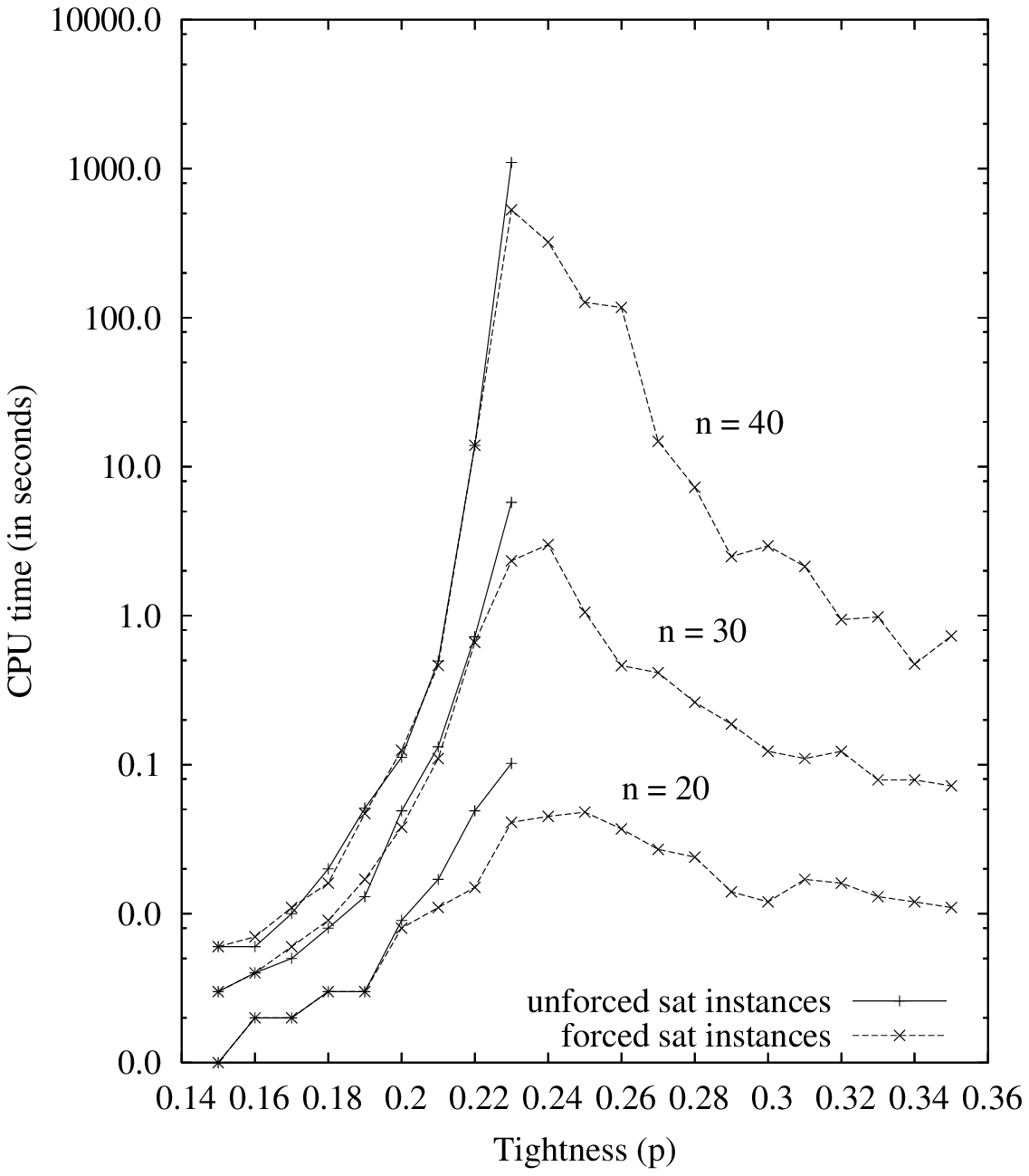}}
\caption{Median search cost ($50$ instances) of solving instances in RB($2$,$\{20,30,40\}$,$0.8$,$3$,$p$) using a tabu search} %Observed phase transitions when $\alpha=0.8$, $r=3$, $k=2$ and $n \in \{20,30,40\}$}
\label{fig:tabu}
\end{figure}

Finally, Figure \ref{fig:tabu} shows the results obtained with a tabu search with respect to the binary instances that have been previously considered with MAC (see Figure \ref{fig:phase1}).
The search effort is given by a median cost since when using an incomplete method, there is absolutely no guarantee of finding a solution in a given limit of time.
Remark that all unsatisfiable (unforced) instances below the threshold have been filtered out in order to make a fair comparison.  
%Results that have been obtained by using an incomplete method are similar to those obtained by MAC:   
It appears that both complete and incomplete methods behave similarly.
In Figure \ref{fig:tabu}, one can see that the search effort grows exponentially with $n$ and that forced instances are as hard as unforced ones.

\section{Related work \label{sec:related}}

%In this paper, we have tried to emphasize the nice theoretical and practical properties of models RB.
%But, as already mentioned in the introduction, other models of random CSP instances exist \cite{A_random,G_random,S_constructing,M_models,GC_consistency}.
%However, we believe that all these models do not offer a framework which is as simple as the one proposed by model RB.
%Nevertheless, we have to point out that the generalized flawless model of \cite{GC_consistency}, although requiring the enforcement of a form of local consistency, allows generating instances with an exponential resolution complexity no matter how large the constraint tightness is.

As a related work, we can mention the recent progress on generating hard satisfiable SAT instances.
\cite{B_hiding,JMS_from} have proposed to build random satisfiable 3-SAT instances on the basis of a spin-glass model from statistical physics.
%Instances are generated while using different probabilities in order to select clauses with $0$, $1$ or $2$ negative literals.
Another approach, quite easy to implement, has also been proposed by \cite{A_hiding}: any 3-SAT instance is forced to be satisfiable by forbidding the clauses violated by both an assignment and its complement. 

Finally, let us mention \cite{A_generating} which propose to build random instances with a specific structure, namely, instances of the
Quasigroup With Holes (QWH) problem.
The hardest instances belong to a new type of phase transition, defined from the number of holes, and coincide with the size of the backbone.
 
\section{Conclusion}

In this paper, we have shown, both theoretically and practically, that the models RB (and RD) can be used to produce, very easily, hard random instances. %, i.e., instances which admit an exponential resolution complexity.
More importantly, the same result holds for instances that are forced to be satisfiable.
To perform our experimentation, we have used some of the most efficient complete and incomplete CSP solvers.
We have also encoded some forced binary CSP instances of class RB($2$,$n$,$0.8$,$0.8/ln\frac{4}{3}$,$p = p_{cr} = 0.25$) with $n$ ranging from $40$ to $59$ into SAT ones (using the direct encoding method) and submitted them to the SAT competition 2004\footnote{\url{http://www.nlsde.buaa.edu.cn/~kexu/benchmarks/benchmarks.htm}}.
About $50\%$ of the competing solvers have succeeded in solving the SAT instances corresponding to $n=40$ ($d=19$ and $m=410$) whereas only one solver has been successful for $n=50$ ($d=23$ and $m=544$). 

Although there are some other ways to generate hard satisfiable instances, e.g. QWH \cite{A_generating} or 2-hidden \cite{A_hiding} instances, we think that the simple and natural method presented in this paper, based on models with exact phase transitions and many hard instances, should be well worth further investigation.  

\section*{Acknowledgments}
The first author is partially supported by NSFC Grant 60403003 and FANEDD Grant 200241. 
Other authors  have been supported by the CNRS, the ``programme COCOA de la R\'egion Nord/Pas-de-Calais'' and by the ``IUT de Lens''.

\appendix

%\bibliography{../globalBiblio}
%\bibliographystyle{named}

\end{document}